%% file: root.tex
\documentclass[letterpaper, 10 pt, conference]{ieeeconf}  
\IEEEoverridecommandlockouts                              
\usepackage{graphics} 
\usepackage[export]{adjustbox}  
\usepackage{times} 
\usepackage{cite}
\usepackage{amsmath,amssymb}
 
\usepackage{amsthm}  

\usepackage{multirow}
\usepackage{xspace}
\usepackage{booktabs}  

\newtheorem{defn}{Definition}

\makeatletter
\DeclareRobustCommand\onedot{\futurelet\@let@token\@onedot}

\makeatother

\usepackage{xcolor}
\definecolor{MyLinkColor}{rgb}{0,0.07843,0.45098}
\usepackage[pdftex,colorlinks=true,allcolors=MyLinkColor]{hyperref}
\title{\LARGE \bf
Active Contact Sensing for Robust Robot-to-Human Object Handover
}

\author{
Linfeng Li,
Lin Shao
and David Hsu
}

\input{macros}
\begin{document}

\maketitle
\thispagestyle{empty}
\pagestyle{empty}

\begin{abstract}
Robot-to-human object handover is an essential skill for robot assistants, from serving drinks at home to passing surgical tools in the operating room. We expect robots to perform handover robustly---to release the object only after a firm human grasp while ignoring incidental touches. Existing passive-sensing methods struggle to generalize across diverse objects and human behaviors, as they lack informative perturbations to disambiguate different contact conditions, such as firm grasp versus incidental touch. We propose an active sensing approach for robust handovers: the robot applies information-gathering motions and senses the resulting human-applied forces to infer the contact state. A firm grasp produces forces in multiple directions, while an accidental touch does not. To capture this distinction, we model the contact state with a Bayesian linear model: a distribution over piecewise-linear mappings from robot motions to human-applied forces. This model enables firm grasp detection and active information gathering. In experiments with 12 participants and 30 diverse rigid objects, our method achieved a 97.5\% success rate---over 30\% higher than two common baselines.
\end{abstract}


\section{Introduction}

Robot-to-human object handover is a fundamental skill for robot helpers, enabling applications from passing surgical tools in the operating room to delivering drinks in daily life~\cite{koene2014experimental, wagner2024robotic, bohren2011towards}. Physical interaction during handover starts when the human touches the object and ends when the robot releases it. To ensure safety and fluency, the robot must release only after the human has formed a firm grasp, while ignoring incidental touches (\figref{fig:teaser}). In practice, this decision often relies on signals from a wrist-mounted force-torque (FT) sensor, as contact forces provide direct information on the human-object interaction. However, interpreting these signals is difficult: both objects and human behaviors vary widely, and heuristics that work for one combination often fail for another. For example, a common approach triggers release when the object is pulled upward~\cite{chan2013human,medina2016human,kupcsik2018learning}, yet an accidental touch from below produces the same sensor signal. This ambiguity reflects a deeper issue: different contact conditions produce distinct interaction patterns, but passive-sensing methods lack informative perturbations to distinguish them; to compensate, they assume deliberate human actions (e.g., an upward pull)---assumptions that do not generalize.

\input{segments/teaser}

We take an active-sensing approach for robust handovers: By applying small probing motions and measuring the resulting human-object interaction forces with the FT sensor, the robot infers the human-object contact condition. Across diverse object types and human behaviors, a firm human grasp resists probing in multiple directions, while an accidental touch does not. However, the force responses are noisy, and the robot's probing motions must respect its dynamics. The robot needs a representation that extracts information robustly from motion-force data and guides feasible probing.

We represent the hand-object contact state as a distribution over piecewise-linear mappings from desired robot motions to human-applied forces. The piecewise-linear structure aligns with the non-smooth unilateral nature of contact, providing an efficient approximation of the force range for firm-grasp detection. The probabilistic formulation quantifies uncertainty, which both ensures robust detection under uncertainty and guides active-sensing motions.

In implementation, we model the contact state as a Bayesian linear model, updated iteratively online using the robot's historical action-observation data. A grasp is deemed firm if, at a chosen confidence level, the model predicts the human can both support the object's weight and apply opposing forces; these conditions are verified via robust linear programs. To decide active-sensing motions, we combine the model with the robot's dynamics and optimize feasible motion commands to minimize model uncertainty.

We conducted handover experiments with 12 human receivers using 30 rigid objects, selected based on the YCB dataset~\cite{calli2015ycb} and a prior human study~\cite{choi2009list} (\figref{fig:objects}). We compared our method against two baselines: one that releases the gripper when the sensed object load diminishes, and another that releases when the force reading changes beyond a threshold. Our proposed method achieved a 97.5\% success rate---over 30\% higher than the two baselines.

Our proposed method only utilizes force feedback and is limited to rigid objects. Future work could explore incorporating additional modalities, such as vision and verbal cues, and extending the approach to deformable objects.

\section{Related Work}

Human-robot object handover is a joint action between a giver and a receiver consisting of two phases: a \textit{pre-handover} phase, in which the partners coordinate their intentions and align their poses for transfer, and a \textit{physical handover} or \textit{transfer} phase~\cite{ortenzi2021object_handovers, strabala2013toward}, during which the receiver grasps the object and the giver releases it.
This work focuses on the \textit{physical handover} phase of \textit{robot-to-human} handover, where the robot serves as the giver and the human as the receiver. In this phase, the robot should release the object only after the human establishes a firm grasp while ignoring incidental touches.

Many methods use heuristics to decide the gripper release, such as releasing when changes in force, displacement, or voltage readings exceed a threshold~\cite{deyle2010rfid, bohren2011towards, eguiluz2017reliable, han2019effects}, monitoring the robot's grasp stability and releasing upon loss \cite{nagata1998delivery}, or following predefined human action protocols \cite{he2022bidirectional}. Others learn a handover controller from human data~\cite{chan2013human,medina2016human,kupcsik2018learning,ferrari2024compliant}, sometimes with additional sensors~\cite{parastegari2016fail}, or learn release conditions directly from raw signals~\cite{grigore2013joint,khanna2022human}. However, these passive-sensing methods observe without informative perturbations and thus cannot disambiguate different contact conditions. To compensate, they assume deliberate human actions---specified a priori or implicitly encoded in the training data (e.g., an upward pull)---which limits generalization. In contrast, we actively sense how the robot's motions produce force responses, enabling firm-grasp detection that generalizes across diverse objects and human behaviors.

This active-sensing capability builds on a probabilistic contact state model that predicts how human-applied forces respond to small robot motions, capturing the motion-force relationship needed to infer contact conditions and decide informative probing actions. Previous work commonly models contact either as continuous impedance relations between motion and force~\cite{love1995environment, erickson2003contact, diolaiti2005contact, haddadi2008new, rozo2016learning, roveda2022sensorless} or as discrete symbolic states in hidden Markov models~\cite{grigore2013joint}. The former assumes fixed contact modes, and the latter simplifies the interaction into a few high-level states; both fail to capture the variability of human grasping behaviors and do not support active sensing. By incorporating suitable features within a Bayesian linear regression framework, our model extends the classic linear impedance model into a probabilistic piecewise-linear model that generalizes across diverse objects and human behaviors while enabling active sensing.

\section{Overview}
\label{sec:problem_formulation}

We address the problem of handing over rigid objects from a robot to a human. The robot starts by holding the object and measuring its positive-valued weight $\objectLoad \in \R_{>0}$ in Newtons with a wrist-mounted force-torque (FT) sensor. Then, our method takes as inputs the robot's proprioception, the interaction forces between the human and the object $\interactionForce \in \realVectorSpace{3}$ inferred from FT sensor measurements~(\appref{app:momentum-observer}), and an approximate position of the human hand \(\humanHandPos \in \R^3 \) detected by an RGB-D camera; it outputs the binary gripper release command and the desired gripper position \( \referencePositionD \in \realVectorSpace{3}\) to hand over the object to the human. From $\referencePositionD$, a compliant controller continuously computes and tracks a desired linear velocity $\robotAction\in\realVectorSpace{3}$ while keeping the gripper orientation fixed~\cite{kronander2016passive}.

In robot-to-human handovers, passive force measurements are often ambiguous. As shown in \figref{fig:blr-example}(a)--(b), an incidental touch and a firm grasp can produce indistinguishable force signals. Resolving this ambiguity requires active sensing, where the robot applies small probing motions to induce force responses that reveal how the human is interacting with the object. \figref{fig:blr-example}(c)--(d) demonstrate how active sensing induces distinctive data patterns. To robustly interpret these force responses and to guide active sensing, we need a representation of the human-object contact state.

We represent this contact state as the relationship between the robot's intended motion~$\robotAction$ and the resulting human-object interaction force~$\interactionForce$. This choice is motivated by the physics of contact: interaction forces arise only when an intended motion is resisted by contact. In our setting, the human hand generates contact forces on the object that resist the robot's intended gripper motion. Consequently, the motion-force relationship provides a direct representation of whether, and in which directions, the human resists the object's motion. This perspective connects to classic linear impedance models, which can be viewed as a linear approximation of this motion-force relationship~\cite{love1995environment,erickson2003contact,diolaiti2005contact,haddadi2008new,rozo2016learning,roveda2022sensorless}.

We model the contact state~$\interactionModel$ as a distribution over piecewise-linear mappings from desired velocity $\robotAction$ to interaction force $\interactionForce$, implemented using a Bayesian linear model (\secref{sec:interaction-model}). The piecewise-linear structure reflects the unilateral, non-smooth nature of contact; the probabilistic formulation accounts for sensor noise and quantifies the uncertainty arising from insufficient probing. We leverage this probabilistic piecewise-linear structure to determine if the human grasp is firm and to plan active sensing actions. To determine grasp firmness, we check if the estimated contact state predicts that the human can both support the object's weight and apply opposing forces (\secref{sec:detecting-firm-grasps}). To plan active sensing actions, we leverage the model's uncertainty: by predicting how candidate motions reduce future uncertainty, we select and execute feasible motions that maximize information gain (\secref{sec:planning}). The overall pipeline of our system is illustrated in \figref{fig:overview}.

\input{segments/overview}

\section{A Bayesian Linear Model of Contact}
\label{sec:interaction-model}

\input{segments/model-example.tex}

In this section, we describe how we model the hand-object contact state $\interactionModel$ as a distribution over piecewise-linear mappings between the robot's desired motion $\robotAction$ and the human-applied forces $\interactionForce$ using Bayesian linear regression.

\subsection{Bayesian Linear Regression}

We now revisit the basics of Bayesian linear regression. Consider $\historyLen$ scalar input-output pairs
$\dataset = \{\tuple{\robotActionScalar_{\bufferIndex}, \interactionForceScalar_{\bufferIndex}}\}_{\bufferIndex=1}^{\historyLen}$,
Bayesian linear regression models the relationship between the input $\robotActionScalar_{\bufferIndex}$ and the output $\interactionForceScalar_{\bufferIndex}$ as
\begin{equation}
  \interactionForceScalar_{\bufferIndex} = \blrWeight^{\top} \feature(\robotActionScalar_{\bufferIndex}) + \noise,~
  \noise \sim \gaussianDistribution\left(0, \blrVariance \identityMatrix\right),
  \label{eqt:bayesian-linear-model}
\end{equation}
where $\blrWeight\in \R^{\blrWeightDimension}$ is the weight vector, $\noise$ is the measurement noise from a Gaussian distribution with zero mean and covariance $\blrVariance \identityMatrix$, $\blrVariance\in R_{>0}$ is a known constant, $\identityMatrix$ is the identity matrix of appropriate size, and $\feature: \R \rightarrow \R^{\blrWeightDimension}$ is a constant feature mapping~\cite{bishop2006pattern}.

Bayesian linear regression treats the weight vector $\blrWeight$ as a random variable with a Gaussian prior distribution $\gaussianDistribution\left(\blrWeightMean_{0}, \blrWeightCovariance_{0}\right)$. Given the input-output pairs $D$, the posterior distribution of the weight vector $\blrWeight$ is also Gaussian, with mean $\blrWeightMean$ and covariance $\blrWeightCovariance$ given by
\begin{equation}
  \begin{split}
    \blrWeightMean &= \blrWeightCovariance \left( \blrWeightCovariance_{0}^{-1} \blrWeightMean_{0} +
                     \blrVariance^{-1} \sum_{\tuple{\robotActionScalar, \interactionForceScalar}\in\dataset}
                     \interactionForceScalar \feature(\robotActionScalar) \right), \\
    \blrWeightCovariance^{-1} &= \blrWeightCovariance_{0}^{-1} +
                                \blrVariance^{-1} \sum_{\tuple{\robotActionScalar, \cdot}\in\dataset}
                                \feature(\robotActionScalar) \feature(\robotActionScalar)^{\top}.
  \end{split}
  \label{eqt:blr-weight-posterior}
\end{equation}

We represent the contact state model using the mean and covariance of the Gaussian distribution $\interactionModel = \tuple{\blrWeightMean, \blrWeightCovariance}$. For a new input $\robotActionScalar'$, the output $\interactionForceScalar'$ predicted by this model is also a Gaussian distribution
\begin{equation}
  \interactionForceScalar' \sim \gaussianDistribution\left(
    \blrWeightMean^{\top} \feature(\robotActionScalar'),
    \feature(\robotActionScalar')^{\top}\blrWeightCovariance\feature(\robotActionScalar')
  \right).
  \label{eqt:blr-predictive-distribution}
\end{equation}

\subsection{Hand-Object Contact State Modeling}

In this subsection, we present our choice of the inputs, outputs, and feature mapping for the Bayesian linear regression model that represents the hand-object interactions.

We select the corresponding directional components from the desired velocity $\robotAction_{\timestep}$ and interaction force $\interactionForce_{\timestep}$ at timestep $\timestep$,
\begin{align*}
\robotAction_{\timestep} = \left[ \robotActionScalarX_{\timestep}\; \robotActionScalarY_{\timestep}\; \robotActionScalarZ_{\timestep}\right]^{\top} \text{ and } \interactionForce_{\timestep} = \left[ \interactionForceScalarX_{\timestep}\; \interactionForceScalarY_{\timestep}\; \interactionForceScalarZ_{\timestep}\right]^{\top},
\end{align*}
as the input and output of the Bayesian linear regression model. In this work, we choose $\robotActionScalar = \robotActionScalar^{z}_{\timestep}$ as the input and $\interactionForceScalar = \interactionForceScalar^{z}_{\timestep}$ as the output, and the model predicts the human reaction to the robot's vertical motion. We only consider the vertical direction as it provides reasonably sufficient information to determine whether the human can support the object's weight and apply opposing forces. This design choice was guided by observations from a pilot study involving four participants, which showed that vertical direction alone was a reliable indicator of grasp stability. Admittedly, it can fail when the human receiver applies a strong lateral push, where friction may be misinterpreted as opposing normal forces.

 We choose the feature mapping as
 \begin{equation}
  \feature(\robotActionScalar) = \left[ \relu(\robotActionScalar)\;\;   -\relu(-\robotActionScalar) \right]^{\top},
  \label{eqt:blr-feature-mapping}
\end{equation}
where $\relu(\cdot)$ is the rectified linear unit function
\begin{align*}
  \relu(s) =
  \left\{\begin{array}{ll}
    s & \text{if } s \geq 0, \\
    0 & \text{otherwise}.
  \end{array}\right.
\end{align*}

With the above choices of input, output, and features, the Bayesian linear regression model in~\eqref{eqt:blr-predictive-distribution} can be interpreted as a piecewise-linear model with dimension $\blrWeightDimension = 2$, consisting of two segments: one active when the robot moves upward, and the other when it moves downward. In each of the segments, only half of the elements in the weight $\blrWeight$ are active, and the active elements can be interpreted as linear impedances. This mirrors the unilateral nature of contact: the human-object interaction can be viewed as multiple non-sticking contacts, each resisting motion along a single direction, whose positive span produces the overall contact effect~\cite{lynch2017modern}.

\subsection{Implementation Details}

The contact state model~$\interactionModel_{\timestep}$ is updated at 200 Hz using the most recent 200 pairs of desired velocities~$\robotAction$ and interaction forces~$\interactionForce$. At each timestep~$\timestep$, the newest pair~$\tuple{\robotAction_{\timestep}, \interactionForce_{\timestep}}$ is appended to the data buffer~$\dataset$, and the oldest sample is removed once the buffer is full; the updated~$\dataset$ is then used to compute the new model~$\interactionModel_{\timestep}$ according to~\eqref{eqt:blr-weight-posterior}. In implementation, this update is performed recursively as described in~\appref{app:recursive-blr}.

\subsection{Examples}
\label{sec:blr-example}

\figref{fig:blr-example} illustrates several examples of the regressed contact-state models under different combinations of human-object contact and robot motion. \figref{fig:blr-example}(a)--(b) show a passive sensing scenario where robot motion $\robotActionScalar$ is only negative (downward).
Because the robot never excites the positive direction, the model has low variance only for $\robotActionScalar < 0$, while the segment for $\robotActionScalar > 0$ remains uncertain. This uncertainty reflects the model's inability to distinguish a light touch from a firm grasp when only downward robot motion is observed.

In contrast, \figref{fig:blr-example}(c)--(d) demonstrate the active sensing scenario, where probing motions span both negative and positive directions. This bidirectional excitation substantially reduces uncertainty across both segments. As a result, the motion-force relationships for a light touch (c) and a firm grasp (d) become clearly different: the firm grasp produces consistent force responses in both directions, whereas the light touch does not.

These examples demonstrate how our contact-state model captures both the motion-force relationship and its associated uncertainty. The uncertainty allows us to robustly detect firm grasps and to decide informative probing motions.

\section{Evaluating Grasp Firmness}
\label{sec:detecting-firm-grasps}

In this section, we describe how to decide whether the human has firmly grasped the object given a contact state model $\interactionModel$. The main idea is to check if the human can support the object's weight and apply opposing forces, as indicated by the model's feasible outputs. If a firm grasp is detected, we open the robot gripper to release the object.

\subsection{Firm Grasp as Feasibility of Desired Outputs}

We define a ``firm grasp'' as the human's ability to exert a specified set of target forces on the object. Since~\eqref{eqt:blr-predictive-distribution} models the relationship between the robot's desired velocity and the resulting human-applied interaction forces, we evaluate whether a given target force $\targetInteractionForceS$ is feasible under the model—i.e., whether it can be produced by the model for some $\robotActionScalar$ within a specified range. This range is defined as $\robotActionSpace = \left\{u \mid -\maxVelocity \leq u \leq \maxVelocity\right\}$, based on the velocity bound $\maxVelocity$ introduced in~\eqref{eqt:velocity-dynamics}. Because the model outputs a distribution rather than a single value, we formalize this notion of feasibility in probabilistic terms.

\begin{defn}[$\confidence$-feasibility]
  Given a model $\interactionModel=\tuple{\blrWeightMean, \blrWeightCovariance}$, we say that the target force $\targetInteractionForceS$ is $\confidence$-feasible if
  \begin{equation}
    \targetInteractionForceS \in \outputSet\left(\blrWeight, \robotActionSpace\right)
    :=\left\{\left.
        \blrWeight^{\top} \feature(\robotActionScalar)
      \right| \robotActionScalar \in \robotActionSpace
    \right\}
    \label{eqt:full-c-feasibility}
  \end{equation}
  for all $\blrWeight \in \blrWeightSet_{\confidence} $, where $\blrWeightSet_{\confidence}$ is the $\confidence$ confidence ellipsoid of the Gaussian distribution with mean $ \blrWeightMean$ and covariance $\blrWeightCovariance$
  \begin{equation}
    \blrWeightSet_{\confidence}= \left\{
      \blrWeightMean + \confidenceEllipsoidMatrix \aVector
    \left|
      \norm{\aVector} \leq 1,~ \aVector \in \R^{2}
    \right.\right\},
  \label{eqt:blr-weight-set-confidence}
 \end{equation}
where $\confidenceEllipsoidMatrix$ is a positive-definite matrix calculated from the confidence level $0<\confidence<1$ and the covariance $\blrWeightCovariance$.
  \label{def:c-feasibility}
\end{defn}

In this work, we consider the minimal set of target forces
\begin{equation}
  \targetInteractionForceSet = \left\{
    0.5 \objectLoad, 0.5, -0.5
  \right\}.
\end{equation}
The first element is set to half of the object's weight, reflecting the assumption that the human can support the object. The second and third elements represent a pair of opposing forces, indicating the ability to reject perturbations. We selected $0.5$ N because it exceeds the sensor's typical noise floor while remaining small enough to ensure a smooth handover. These values were validated through a pilot study with four participants.

Given a confidence level $\confidence$, we consider a grasp to be firm if all target forces in the set $\targetInteractionForceSet$ are $\confidence$-feasible. In this work, we set $\confidence = 0.99$ to ensure a high level of robustness in verifying the firmness of the human grasp.

\subsection{Checking Output Feasibility}

Having defined $\confidence$-feasibility as the criterion for a firm grasp, we now describe how to verify whether a target force $\targetInteractionForceS$ satisfies this condition. In this subsection, we present a sufficient condition for $\targetInteractionForceS$ to be $\confidence$-feasible under a given model $\interactionModel= \tuple{\blrWeightMean, \blrWeightCovariance}$. This verification is formulated and implemented as robust linear programs~\cite{ben-tal2009robust_optimization}.

Since the model is piecewise linear, we partition the model's feasible input $\robotActionSpace$ into 2 pieces, in each piece the model is linear. We denote the two sets of input as $\robotActionSpace^{-}=\{\robotActionScalar | -\maxVelocity \leq \robotActionScalar \leq 0 \}$ and $\robotActionSpace^{+} = \{\robotActionScalar | 0 \leq \robotActionScalar \leq \maxVelocity \}$, where $\maxVelocity \in \mathbb{R}{>0}$ is the maximum gripper velocity.

Let $\robotActionSpace'$ denote either of the two partitions, $\robotActionSpace^{-}$ or $\robotActionSpace^{+}$. Since $\robotActionSpace' \subset \robotActionSpace$, a target force $\targetInteractionForceS$ is $\confidence$-feasible if it is in the corresponding output set $\outputSet\left(\blrWeight, \robotActionSpace'\right) \subset \outputSet\left(\blrWeight, \robotActionSpace\right)$ for all $\blrWeight \in \blrWeightSet_{\confidence}$. Thus, we can check $\confidence$-feasibility separately for each partition: as long as the target force is $\confidence$-feasible within at least one partition, it satisfies the overall check.

Our feature $\feature$ defined in~\eqref{eqt:blr-feature-mapping} selects the component of the weight vector $\blrWeight$ corresponding to the sign of the input. Therefore, for an input partition $\robotActionSpace' \in \{ \robotActionSpace^{+}, \robotActionSpace^{-} \}$, we can select the corresponding dimension of the weight vector $\blrWeight$ as $\blrWeightS' = \selectionMatrix'\blrWeight$ such that $w' u = \blrWeight^{\top} \feature(\robotActionScalar)$, where $\selectionMatrix^{+} = \left[1~ 0 \right]$  and $\selectionMatrix^{-} = \left[0~ 1 \right]$ are the selection matrices. Then, we can rewrite the output set in~\eqref{eqt:full-c-feasibility} over $\robotActionSpace'$ as
\begin{equation}
  \outputSet\left(\blrWeight, \robotActionSpace'\right) =
  \left\{ \left. \blrWeightS' \robotActionScalar \right| \robotActionScalar \in \robotActionSpace',~ w' = \selectionMatrix' \blrWeight \right\},
  \label{eqt:partition-output-set}
\end{equation}
which is a convex set because linear transformation preserves convexity of $\robotActionSpace'$~\cite{boyd2004convex_optimization}.

Taking advantage of the convexity within the partition $\robotActionSpace'$, we propose a sufficient condition for $\confidence$-feasibility as the feasibility of two robust linear programs (LPs)
\begin{equation}
\begin{array}{cl}
\min_{\robotActionScalar} & 0\cdot \robotActionScalar \\
\text{s.t.}
& u\in \robotActionSpace' \\
& \left(\selectionMatrix' \left(\blrWeightMean + \confidenceEllipsoidMatrix \aVector\right)\right) \robotActionScalar \leq \targetInteractionForceS ~\forall~ \norm{\aVector} \leq 1
\end{array}
\label{eqt:robust-lp-less-than-target}
\end{equation}
and
\begin{equation}
\begin{array}{cl}
\min_{\robotActionScalar} & 0\cdot \robotActionScalar \\
\text{s.t.}
& u\in \robotActionSpace' \\
& \left(\selectionMatrix' \left(\blrWeightMean + \confidenceEllipsoidMatrix \aVector\right)\right) \robotActionScalar \geq \targetInteractionForceS ~\forall~ \norm{\aVector} \leq 1.
\end{array}
\label{eqt:robust-lp-greater-than-target}
\end{equation}
If the first robust LP~\eqref{eqt:robust-lp-less-than-target} is feasible, there exists $\interactionForceScalar^{\text{less}}\in\outputSet\left(\blrWeight, \robotActionSpace'\right)$ such that $\interactionForceScalar^{\text{less}}\leq \targetInteractionForceS$ for all $\blrWeight \in \blrWeightSet_{c}$; similarly, the second~\eqref{eqt:robust-lp-greater-than-target} checks $\interactionForceScalar^{\text{greater}}\geq \targetInteractionForceS$. If both robust LPs are feasible, we have
\begin{equation}
  \interactionForceScalar^{\text{less}}, \interactionForceScalar^{\text{greater}}
  \in \outputSet\left(\blrWeight, \robotActionSpace'\right)
  \text{ for all } \blrWeight \in \blrWeightSet_{c}.
\end{equation}
Since $\interactionForceScalar^{\text{less}}\leq \targetInteractionForceS \leq \interactionForceScalar^{\text{greater}}$, and  $\outputSet\left(\blrWeight, \robotActionSpace'\right)$ is convex, we have $\targetInteractionForceS\in \outputSet\left(\blrWeight, \robotActionSpace'\right) \subset\outputSet\left(\blrWeight, \robotActionSpace\right)$ for all $\blrWeight \in \blrWeightSet_{c}$, which is \defref{def:c-feasibility}.

The two robust LPs are equivalent to two second-order cone programs (SOCPs)~\cite{ben-tal2009robust_optimization}, which can be efficiently solved using off-the-shelf solvers.

In summary, given an estimated model $\interactionModel$, we consider the human grasp to be firm if all target forces in the set $\targetInteractionForceSet$ are $\confidence$-feasible. To check the feasibility of a target force $\targetInteractionForceS\in\targetInteractionForceSet$, we solve the two robust LPs~\eqref{eqt:robust-lp-less-than-target} and~\eqref{eqt:robust-lp-greater-than-target} for each partition $\robotActionSpace'\in \left\{\robotActionSpace^{+}, \robotActionSpace^{-}\right\}$; the target force is $\confidence$-feasible if both robust LPs are feasible in any partition. 

In implementation, firm-grasp detection runs at 50 Hz. At each timestep~$\timestep$, we evaluate the latest contact-state model~$\interactionModel_{\timestep}$; if a firm grasp is detected, the robot releases the object, otherwise it continues holding it.

\section{Deciding Informative Motion}
\label{sec:planning}

As shown in \secref{sec:blr-example}, without informative motions $\robotAction_{\timestep}$, the contact state model retains high uncertainty, making it difficult to distinguish a firm grasp from a light touch. To resolve such ambiguity, we select robot motions that both reduce contact state uncertainty and respect the robot's dynamics.

To formally reason about which motions are informative, we first quantify the uncertainty of the contact state model~$\interactionModel$ using the entropy of its weight distribution $\blrWeight \sim \gaussianDistribution(\blrWeightMean, \blrWeightCovariance)$~\cite{thrun2005probabilistic_robotics}. The entropy is $\entropy(\blrWeight) = 0.5\blrWeightDimension (1 + \log 2\pi) + 0.5 \log \determinant{\blrWeightCovariance}$, where $\determinant{\,\cdot\,}$ denotes the matrix determinant. Since the entropy of a Gaussian depends only on its covariance, we write $\entropy(\blrWeightCovariance)$ to denote the entropy of the contact state model. Given two models $\interactionModel' = \tuple{\blrWeightMean', \blrWeightCovariance'}$ and $\interactionModel = \tuple{\blrWeightMean, \blrWeightCovariance}$, the information gain from the former over the latter is defined as
\begin{equation}
\informationGain(\blrWeightCovariance, \blrWeightCovariance')
= \entropy(\blrWeightCovariance) - \entropy(\blrWeightCovariance').
\label{eqt:information-gain}
\end{equation}
A larger information gain corresponds to a greater reduction in model uncertainty.

To evaluate the information gain for potential probing motions, we must predict how the robot's future actions will affect the model's covariance.
The covariance $\blrWeightCovariance$ only depends on the inputs $\robotAction$ in the data~$\dataset$ of the model~\eqref{eqt:blr-weight-posterior}. Hence, if we can predict future $\robotAction$ resulting from a $\referencePositionD$, we can predict the information gain obtained by applying $\referencePositionD$.

The desired velocity $\robotAction$ is computed from the desired reference position $\referencePositionD$ and the actual position $\robotPosition$. Specifically,
\begin{equation}
  \robotAction = \left\{
    \begin{array}{ll}
      \robotActionRaw & \text{if } \norm{\robotActionRaw} \leq \maxVelocity, \\
      \maxVelocity\robotActionRaw / \norm{\robotActionRaw}  & \text{otherwise,}
    \end{array}
  \right.
  \label{eqt:velocity-dynamics}
\end{equation}
where $\maxVelocity \in \mathbb{R}{>0}$ is the maximum gripper velocity, $\robotActionRaw \in \mathbb{R}^3$ is the raw desired velocity, and $|\cdot|$ denotes the Euclidean norm. The raw velocity is given as $\robotActionRaw = \DSGain \identityMatrix (\referencePosition - \robotPosition)$, where $\DSGain \in \mathbb{R}{>0}$ is a scalar gain, and $\referencePosition \in \mathbb{R}^3$ is the smoothened desired reference position $\referencePositionD \in \mathbb{R}^3$ using the first-order low-pass filter
\begin{equation}
\dot{\referencePosition} = \smoothFilterParam (\referencePositionD - \referencePosition),
\label{eqt:reference-dynamics}
\end{equation}
with $\smoothFilterParam \in \mathbb{R}_{>0}$ set to achieve a cutoff frequency of 10 Hz. This filtering ensures that the commanded motion satisfies the hardware's smoothness constraints.

To predict future $\robotAction$ given $\referencePositionD$, we need to predict future $\robotPosition$. We achieve this through the approximate translational dynamics of the robot gripper
\begin{equation}
  \dot{\robotVelocity} = \robotDamping \left(\robotAction - \robotVelocity \right) +
  \robotMass_{\timestep} ^{-1} \left[
    \begin{array}{c}
      0 \\ 0 \\ \blrWeightMean_{\timestep}^{\top}\feature(\robotActionScalar)
    \end{array}
  \right],~
  \dot{\robotPosition} = \robotVelocity,
  \label{eqt:robot-dynamics}
\end{equation}
where $\robotVelocity$ is the gripper velocity, \( \robotDamping \in \R^{3\times3} \) is a positive-definite constant diagonal damping matrix of the low-level controller, \( \robotMass_{\timestep} \in \R^{3\times 3} \) is the robot gripper translational mass at timestep $t$, treated as a constant. This model has a few simplifications: (i) it ignores the coupling between the translational and rotational motions, (ii) it treates the operational mass as a constant, and (iii) it only has an approximate model of the human forces $\blrWeightMean_{\timestep}^{\top}\feature(\robotActionScalar)$ and treats the model $\blrWeightMean_{\timestep}$ as a constant. Still, this model is sufficiently accurate for our decision-making.

By combining the reference dynamics~\eqref{eqt:reference-dynamics} and the gripper dynamics~\eqref{eqt:robot-dynamics}, we can simulate the closed-loop system and predict how a candidate reference position $\referencePositionD$ influences information gain. Formally, we combine \eqref{eqt:reference-dynamics} and \eqref{eqt:robot-dynamics} as a single differential equation
\begin{equation}
  \dot{\odeState} = \odeModel(\odeState, \referencePositionD),
\end{equation}
where $\odeState$ is the concatenation of $\robotPosition$, $\robotVelocity$, and $\referencePosition$. Given a desired reference position $\referencePositionD$, we use the latest $\robotPosition_{\timestep}$, $\robotVelocity_{\timestep}$, and $\referencePosition_{\timestep}$ as the initial condition, and predict the future state $\odeState$ for a duration of $T$ seconds. In this work, we set $T=0.25$ seconds. Then, we sample the future desired robot velocities $\robotAction$ from the predicted state $\odeState$ using~\eqref{eqt:velocity-dynamics} with the same frequency as the contact state update rate. Using the current data $\dataset_{\timestep}$ and the predicted new input data, we can predict the future model covariance $\blrWeightCovariance'$ using the posterior update~\eqref{eqt:blr-weight-posterior}.
Finally, we predict the information gain~\eqref{eqt:information-gain}.

We select the optimal desired reference position $\referencePositionDStar$ by maximizing the information gain with a regularization term
\begin{equation}
  \referencePositionDStar = \argmax_{\aVector}~
    \informationGain\left(S, \predictCovariance(\aVector)\right) - \planningRegularizationCoeff \norm{\aVector - \robotPosition_{\timestep}},
    \label{eqt:planning-optimization}
\end{equation}
where we use $\predictCovariance(\referencePositionD)$ to denote the prediction of future model covariance $\blrWeightCovariance'$ of a desired reference position~$\aVector$, and $\planningRegularizationCoeff \in \R_{>0}$ is a constant regularization weight. We solve the maximization via grid-search: we sample 11 candidates along the vertical direction around the current gripper position $\robotPosition_{\timestep}$ with a distance range of $\pm 2$ cm.

In implementation, we decide the active-sensing motion at 10 Hz. At each timestep~$\timestep$, we first check whether the human is in contact with the object, defined as~$|\interactionForce_{\timestep}| \ge 0.5~\text{N}$. If no contact is detected, we guide the gripper toward the estimated human hand position~$\humanHandPos_{\timestep}$ by setting~$\referencePositionD_{\timestep} = \humanHandPos_{\timestep}$. If contact is detected, we use the current system state $\odeState_t$ and the estimated contact model~$\interactionModel_{\timestep}$ to compute a new reference position~$\referencePositionD_{\timestep} = \referencePositionDStar$ that generates dynamically feasible probing motions to reduce model uncertainty.

\input{segments/objects}
\input{segments/experiment-setup}

\section{Experiments}

We evaluate our method and two baselines on 30 diverse rigid objects (see \figref{fig:blr-example}) with 12 participants. The primary evaluation criterion is whether the human maintains a firm grasp on the object at the moment the robot releases it. Our method achieves significantly higher success rates than the two baselines, demonstrating its effectiveness in robot-to-human object handovers.

\subsection{Experimental Setup}

We conducted our experiments using a Franka Emika Research 3 (FR3) robotic arm. An ATI Gamma force-torque sensor was mounted on the arm's flange, with a Franka Hand parallel gripper attached to the sensor. To detect the human hand position, we used an Intel RealSense L515 RGB-D camera mounted above the workspace. During handover trials, participants were seated or standing in front of the robot, facing the robot sideways. Participants were blindfolded to evaluate the robot's release decision in isolation, minimizing human compensation for potential robot errors~\cite{ferrari2024compliant}. This also emulates scenarios where humans lack visual feedback---for example, visually impaired users or mechanics working under a vehicle. The overall setup is shown in \figref{fig:experiment-setup}.

We selected 30 objects representing the types defined in the YCB dataset~\cite{calli2015ycb} and a human study~\cite{choi2009list}, excluding non-rigid items and those too heavy or too small. The resulting set spans diverse shapes, sizes, and masses (Fig.~\ref{fig:objects}).

We recruited 12 participants for the study. Each participant interacted with 10 objects sampled from the full set of 30 objects (\figref{fig:objects}); we ensured that each object was tested by four different participants. To minimize learning effects, we randomized the order of objects and methods. Each participant evaluated all three methods sequentially. After completing the 10 trials for each method, they completed a post-experiment survey consisting of two standard questionnaires—NASA Task Load Index without weighting (RTLX)\cite{hart2006nasa} and System Usability Scale (SUS)\cite{brooke1996sus}—as well as an optional section for free-form feedback.

At the beginning of the experiment, participants were instructed to (i) receive the object by grasping it and (ii) avoid moving too quickly or forcefully after grasping. Beyond these guidelines, they were free to interact with the object as they preferred to preserve the diversity of human behaviors. The handover process was demonstrated once before the participant was blindfolded.

Each handover trial began by placing the object in the robot gripper with a predefined orientation. The robot then moved its gripper to a randomly sampled initial pose within a defined range. Participants were told the object type and notified when the trial started. During the trial, if their hand was outside the robot's operational range, we provided verbal cues to guide them back to the robot's operational range.

A handover trial concluded either when the robot released the object---regardless of whether the participant maintained their grasp---or, if the robot failed to release it, when the participant requested intervention to terminate the trial.

Before the experiment, we conducted a pilot study with four participants and 12 objects to determine the parameters of all methods.

\subsection{Baselines}

We compare our method with two baselines. The first baseline, \baselineForce{}, is a simple heuristic that releases the object when the interaction force magnitude $\norm{\interactionForce}$ exceeds 1 Newton. This approach does not take the human's actual grasp status into account and may result in premature releases.

The second baseline, \baselineWeight{}, simplifies a widely used approach in which the robot adjusts its grasping force proportionally to the sensed object load~\cite{chan2013human, medina2016human, kupcsik2018learning}. Because our gripper (Franka Hand) does not support real-time force control and the grasp controller may require object-specific tuning, we adopt a threshold-based release condition instead. Specifically, the robot releases the object if the sensed load drops below 20\% of the object's weight $\objectLoad$—as in~\cite{medina2016human}—or if a non-negligible upward force is detected for lightweight objects. Formally, the object is released if $\interactionForceScalarZ \geq \max\left(0.8\objectLoad, 0.5\right)$. Prior work has shown that this simplification does not reduce the method's success rate~\cite{medina2016human}. This baseline implicitly expects the human to perform a deliberate upward motion to signal readiness for release.

All three methods incorporate visual feedback in the same manner, as described in \secref{sec:planning}. For both baselines, once contact is detected, we set the current position~$\robotPosition$ as the desired reference position~$\referencePositionD$. For example handovers of all methods, please see the supplementary video.

\subsection{Results}

Our method significantly outperforms the two baselines in \emph{success rate}, achieving 97.5\% with low variance, compared to 67.5\% for \baselineForce{} and 65.0\% for \baselineWeight{} (\figref{fig:objective-results}(a)). A handover is considered successful if the object is firmly grasped by the human at the moment of robot release. Grasp firmness is assessed from video recordings by three independent experts blinded to the method, with the final outcome determined by majority vote. We measure our experiment's inter-rater reliability by Fleiss' kappa, which is 0.803, indicating substantial agreement among raters~\cite{landis1977measurement}.

\input{segments/objective-results}
\input{segments/survey-scores}

\baselineForce{} often failed due to premature release, where the robot opened the gripper before the human had firmly grasped the object. It tended to release almost immediately after initial contact, frequently causing the object to drop. Participants' free-form feedback confirmed this issue: ``The system drops the object a bit early'' (Participant \#2, abbreviated as P2), ``...it released the bleach cleanser too quickly for me to catch'' (P3), ``I think this system might be too confident to drop items'' (P5), ``The robot hand released upon the object touching my hand, which is very likely to cause abrupt dropping'' (P10), and  ``Sometimes the arm dropped the item a bit too fast'' (P11).

\baselineWeight{} exhibited two distinct failure modes: premature release of light objects and failure to release heavier ones. When participants applied small but non-negligible upward forces exceeding the object's weight, the robot often released the object before a firm grasp was established. Conversely, when participants held heavier objects without pulling upward strongly, the robot failed to release altogether, leading participants to request intervention and eventually give up. Participants' feedback reflected both types of failures. For light objects, several noted premature releases: ``When handing over toothbrush and pill box, I haven't held objects well, but the robot started to drop them'' (P2), ``...while for some lighter objects it just released very quickly and hard for me to catch'' (P3), and ``The robot sometimes released the object early'' (P12). For heavier objects, participants described difficulty retrieving them: ``I have grabbed the object but the robot is still holding it'' (P2), ``Some objects (usually the ones that are bulky and heavy) cannot be taken from the robot arm even after several attempts''(P3). Because \baselineWeight{} fails due to both premature drops and incomplete handovers, participants found its behavior inconsistent: ``The behavior for the next item is not always similar to the previous item.'' (P6), ``The robot produces less consistent behavior'' (P7), and ``The timing of the robot hand's release is a bit unpredictable'' (P10).

In contrast, our method robustly detected firm grasps, resulting in significantly higher success rates. Most participants found our method reliable: ``This method released the gripper in proper time'' (P1), ``I find it never dropped things early, it always waited until I can steadily hold the object, then released. It seems purposely let the object touch my hand and give me time to find a good pose to hold.'' (P2), and ``The robot performs better on checking whether I get the item'' (P6). Nevertheless, some participants found our method over-cautious: ``...but sometimes it takes too long to check'' (P6).

We conducted two standardized surveys: the Raw NASA Task Load Index (RTLX) and the System Usability Scale (SUS). RTLX scores range from 0 to 100, with lower values indicating lower perceived workload, while SUS scores range from 0 to 100, with higher values indicating greater usability. The survey results for all three methods are summarized in \tabref{tab:survey-scores}. The scores differ by about 10 points on each metric (RTLX: 28.9--39.3; SUS: 56.5--67.9), suggesting comparable workload and usability across methods. Overall, our method achieved the highest success rate while maintaining competitive subjective ratings, demonstrating a strong balance between robustness and user satisfaction.

\section{Conclusions and Discussions}

We have presented an active sensing framework for robot-to-human object handovers, where the robot applies probing motions and measures the resulting force responses to distinguish firm grasps from incidental touches. Central to our approach is a probabilistic piecewise-linear contact model that quantifies uncertainty, enabling us to both robustly detect firm grasp and decide informative probing actions. This active strategy improves robustness, achieving a 97.5\% success rate---over 30\% higher than passive baselines---across 12 participants and 30 diverse rigid objects.

Our results demonstrate that probabilistic, active-sensing strategies substantially improve the robustness of physical human-robot interaction. Nevertheless, our approach is limited to rigid objects and relies solely on force sensing. Additionally, the probing motions, while informative, may slightly prolong the handover process. Future directions include integrating multimodal cues such as vision or language, extending the framework to deformable objects, and developing adaptive release strategies that balance speed and robustness across different application contexts.




\appendix
\subsection{Estimating Human-Object Interaction Force}
\label{app:momentum-observer}

\newcommand{\eeInertia}{\mathbf{I}_{\mathrm{ee}}}
\newcommand{\dotEeInertia}{\dot{\mathbf{I}}_{\mathrm{ee}}}
\newcommand{\eeTwist}{\mathcal{V}}
\newcommand{\eePose}{\mathcal{X}}
\newcommand{\eeCoriolis}{\mathbf{C}_{\mathrm{ee}}}
\newcommand{\eeGravity}{\mathbf{g}_{\mathrm{ee}}}
\newcommand{\eeWrench}{\mathcal{F}}
\newcommand{\eeMomentum}{\mathbf{}{p}}
\newcommand{\eeWrenchFt}{\eeWrench_{\mathrm{FT}}}
\newcommand{\eeWrenchExt}{\eeWrench_{\mathrm{ext}}}
\newcommand{\momentumResidual}{\mathbf{r}}
\newcommand{\observerGain}{\mathbf{K}}

We model the gripper as a single rigid body with dynamics
\begin{align*}
    \eeInertia(\eePose) \dot{\eeTwist} + \eeCoriolis(\eePose, \eeTwist) \eeTwist + \eeGravity(\eePose) = \eeWrenchFt + \eeWrenchExt,
\end{align*}
where $\eePose \in SE(3)$ and $\eeTwist \in \mathbb{R}^6$ are the pose and twist of the gripper,
$\eeInertia$ is the inertia matrix, $\eeCoriolis$ is the Coriolis matrix, $\eeGravity$ is the gravity, $\eeWrenchFt$ is the measured wrench applied to the gripper from the FT sensor's sensing plate, and $\eeWrenchExt$ is the external wrench applied to the gripper.

We estimate the unknown external wrench $\eeWrenchExt$ using a momentum residual observer~\cite{luca2005sensorless}. Defining generalized momentum $\eeMomentum = \eeInertia \eeTwist$, the residual $\momentumResidual$ at time $t$ is given as
\begin{align*}
    \momentumResidual(t) = \observerGain \left[ \int_0^t \left( -\eeCoriolis^{\top}\eeTwist + \eeGravity - \eeWrenchFt - \momentumResidual(s) \right) \mathrm{d}s + \eeMomentum(t) \right].
\end{align*}
Since $\dot{\eeMomentum} = \eeCoriolis^{\top} \eeTwist + \eeWrenchFt + \eeWrenchExt - \eeGravity$, the residual satisfies the first-order dynamics
$\dot{\momentumResidual} = \observerGain(\eeWrenchExt - \momentumResidual)$.
With a positive-definite gain $\observerGain$, the residual $\momentumResidual$ converges to $\eeWrenchExt$, from which we select the translational components as the human-object interaction force $\interactionForce$.

\subsection{Recursive Bayesian Linear Regression}
\label{app:recursive-blr}

Given a model represented by $\tuple{\blrWeightMean, \blrWeightCovariance}$ regressed from the dataset $\dataset$, if an input-output pair $\tuple{\robotActionScalar', \interactionForceScalar'}$ is added to the dataset $\dataset$, the posterior distribution $\blrWeightMean'$ and $\blrWeightCovariance'$ can be recursively updated as
\begin{equation}
  \begin{split}
  \blrWeightMean' &= \blrWeightCovariance' \left( \blrWeightCovariance^{-1} \blrWeightMean + \blrVariance^{-1} \interactionForceScalar' \feature\left(\robotActionScalar'\right) \right), \\
  \left(\blrWeightCovariance'\right)^{-1} &= \blrWeightCovariance^{-1} + \blrVariance^{-1} \feature\left(\robotActionScalar'\right) \feature\left(\robotActionScalar'\right)^{\top}.
  \end{split}
  \label{eqt:blr-weight-posterior-recursive}
\end{equation}
This recursive update is equivalent to modifying the dataset $\dataset^{\prime} = \dataset \cup \{ \tuple{\robotActionScalar', \interactionForceScalar'} \}$, and then perform the posterior update in~\eqref{eqt:blr-weight-posterior} using $\dataset^{\prime}$. Similarly, if we remove the input-output pair $\tuple{\robotActionScalar', \interactionForceScalar'}$ from the dataset $\dataset$, we can update the posterior distribution by changing $+$ in \eqref{eqt:blr-weight-posterior-recursive} to $-$.



\bibliographystyle{IEEEtran} 
\bibliography{references.bib}

\end{document}

%% file: macros.tex
\newcommand{\argmax}{\mathop{\mathrm{arg\,max}}}

\newcommand{\tuple}[1]{\left\langle#1\right\rangle}
\newcommand{\R}{\mathbb{R}}

\newcommand{\realVectorSpace}[1]{\R^{#1}}

\newcommand{\identityMatrix}{\mathbf{I}}
\newcommand{\norm}[1]{\left\|#1\right\|}
\newcommand{\determinant}[1]{\left|#1\right|}

\newcommand{\aVector}{\mathbf{z}}
\newcommand{\relu}{\mathrm{ReLU}}
\newcommand{\entropy}{H}
\newcommand{\informationGain}{IG}
\newcommand{\gaussianDistribution}{\mathrm{Gaussian}}

\newcommand{\robotPosition}{\mathbf{q}}
\newcommand{\robotVelocity}{\mathbf{v}}
\newcommand{\robotDamping}{\mathbf{D}}
\newcommand{\robotActionScalar}{u}
\newcommand{\robotActionScalarX}{u^{\mathrm{x}}}
\newcommand{\robotActionScalarY}{u^{\mathrm{y}}}
\newcommand{\robotActionScalarZ}{u^{\mathrm{z}}}
\newcommand{\robotAction}{\mathbf{\robotActionScalar}}
\newcommand{\robotActionRaw}{\mathbf{\robotActionScalar}^{\mathrm{raw}}}
\newcommand{\robotActionSpace}{\mathbb{U}}
\newcommand{\robotMass}{\mathbf{M}}
\newcommand{\maxVelocity}{v_{\max}}
\newcommand{\DSGain}{k}
\newcommand{\referencePosition}{\mathbf{r}}
\newcommand{\referencePositionD}{\referencePosition^{\mathrm{d}}}
\newcommand{\referencePositionDStar}{\referencePosition^{\mathrm{d\star}}}
\newcommand{\smoothFilterParam}{\alpha}
\newcommand{\interactionForceScalar}{f}
\newcommand{\interactionForceScalarX}{f^{\mathrm{x}}}
\newcommand{\interactionForceScalarY}{f^{\mathrm{y}}}
\newcommand{\interactionForceScalarZ}{f^{\mathrm{z}}}
\newcommand{\interactionForce}{\mathbf{\interactionForceScalar}}

\newcommand{\timestep}{t}
\newcommand{\bufferIndex}{i}
\newcommand{\humanHandPos}{\mathbf{p}}
\newcommand{\historyLen}{N}

\newcommand{\confidence}{c}
\newcommand{\objectLoad}{L}
\newcommand{\odeState}{\mathbf{x}}
\newcommand{\odeModel}{\mathrm{AugmentedDynamics}}
\newcommand{\blrWeightDimension}{d}
\newcommand{\blrWeightS}{w}
\newcommand{\blrWeight}{\mathbf{\blrWeightS}}
\newcommand{\blrWeightSet}{\mathbb{W}}
\newcommand{\blrWeightMean}{\mathbf{m}}
\newcommand{\blrWeightCovariance}{\mathbf{S}}
\newcommand{\blrVariance}{\beta}
\newcommand{\feature}{\boldsymbol{\phi}}
\newcommand{\dataset}{D}
\newcommand{\noise}{\boldsymbol{\eta}}
\newcommand{\confidenceEllipsoidMatrix}{\mathbf{P}}
\newcommand{\selectionMatrix}{\mathbf{E}}
\newcommand{\outputSet}{\mathbb{F}}
\newcommand{\predictCovariance}{\mathrm{pred}}
\newcommand{\planningRegularizationCoeff}{c_{\mathrm{reg}}}
\newcommand{\targetInteractionForceS}{\interactionForceScalar^{\mathrm{target}}}
\newcommand{\targetInteractionForceSet}{\mathbb{F}^{\mathrm{target}}}
\newcommand{\interactionModel}{\mathcal{C}}

\makeatletter
\DeclareRobustCommand\onedot{\futurelet\@let@token\@onedot}
\def\@onedot{\ifx\@let@token.\else.\null\fi\xspace}

\makeatother

\newcommand{\baselineForce}{ForceThr}
\newcommand{\baselineWeight}{WeightThr}

\newcommand{\secref}[1]{Section~\ref{#1}}
\newcommand{\appref}[1]{Appendix~\ref{#1}}
\renewcommand{\eqref}[1]{(\ref{#1})}
\newcommand{\figref}[1]{Fig.~\ref{#1}}

\newcommand{\tabref}[1]{Table~\ref{#1}}
\newcommand{\defref}[1]{Definition~\ref{#1}}

%% file: segments/teaser.tex
\begin{figure}[t]
    \centering
    \includegraphics[scale=0.5]{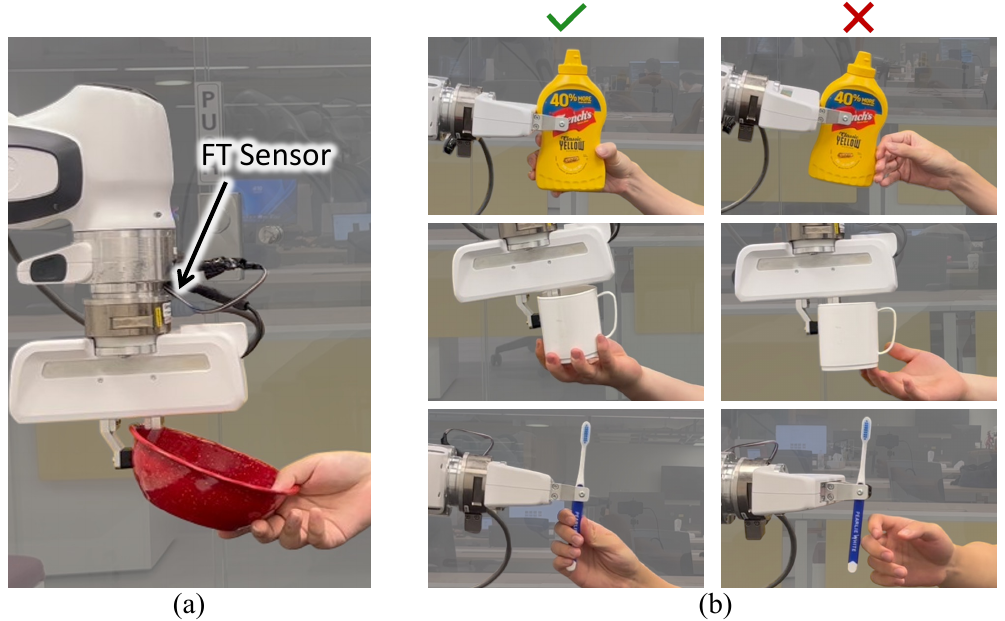}
    \vspace{-0.3cm}
    \caption{(a) We use a wrist-mounted force-torque (FT) sensor to sense human-object contact. (b) Example human-object contacts during robot-to-human handovers: \textcolor{green!50!black}{firm grasps} on the left, where the robot can safely release the object, and \textcolor{red!80!black}{incidental touches} on the right, where premature release risks dropping the object. Across diverse rigid objects and grasp behaviors, our active sensing approach reliably distinguishes between them, while passive methods often fail.}
    \label{fig:teaser}
\end{figure}

%% file: segments/overview.tex
\begin{figure}[t]
    \centering
    \includegraphics[scale=0.5]{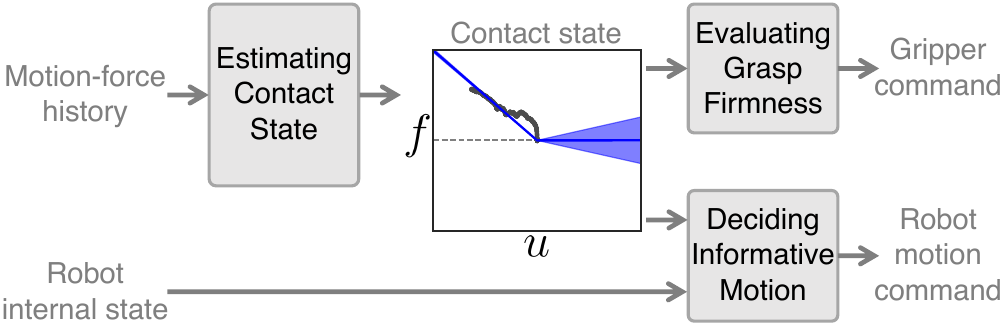}
    \vspace{-0.3cm}
    \caption{Our method centers on a human-object contact state represented as a Bayesian linear model, estimated online from motion-force history (\secref{sec:interaction-model}). The model is used to (i) assess whether the human has formed a firm grasp (\secref{sec:detecting-firm-grasps}) and (ii) plan active-sensing motions that reduce model uncertainty while respecting the robot’s dynamics (\secref{sec:planning}).
    }
    \label{fig:overview}
\end{figure}

%% file: segments/model-example.tex
\begin{figure}[t]
    \centering
    \includegraphics[scale=0.5]{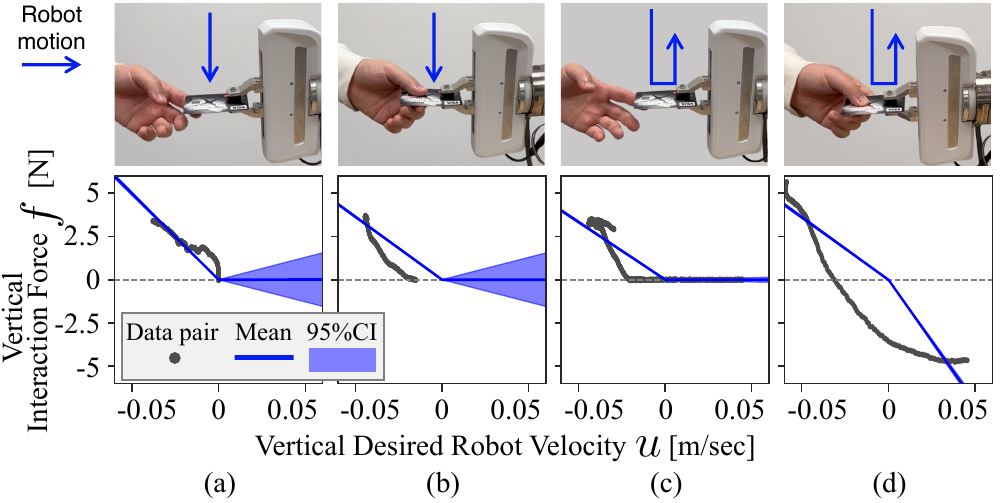}
    \vspace{-0.3cm}
    \caption{
    Contact states modeled as probabilistic piecewise-linear mapping from vertical desired velocity to vertical interaction force. The top row shows the physical contact (blue arrows indicate robot motion). The bottom row shows the estimated model: grey dots are data pairs $\tuple{\robotActionScalar, \interactionForceScalar}$, the blue line is the posterior mean $\blrWeightMean^{\top}\feature(\robotActionScalar)$, and the shaded band is the 95\% confidence interval.
    (a)--(b) Passive sensing scenario: Downward-only motion produces indistinguishable signals for an incidental touch (a) and a firm grasp (b). The model cannot distinguish them, resulting in high uncertainty for the unobserved direction ($u>0$).
    (c)--(d) Active sensing scenario: Bidirectional probing reveals the underlying human-object interaction pattern. The distinct motion-force relationships are estimated with low uncertainty, showing unilateral resistance for an incidental touch (c) versus bilateral resistance for a firm grasp (d).
    }
    \label{fig:blr-example}
\end{figure}

%% file: segments/objects.tex
\begin{figure*}[t]
    \centering
    \includegraphics[scale=0.5]{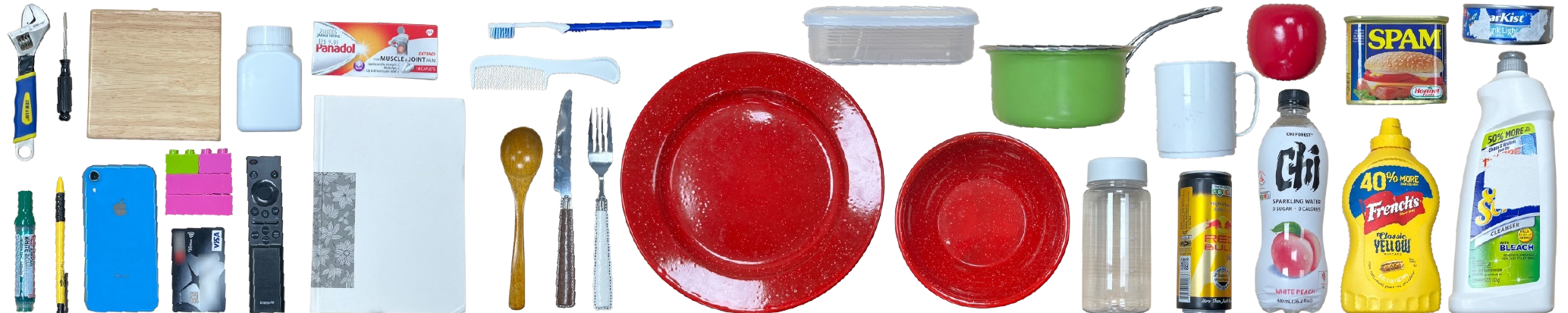}
    \vspace{-0.2cm}
    \caption{We selected 30 types of objects from the YCB dataset~\cite{calli2015ycb} and a human study~\cite{choi2009list}. These objects span a broad range of shapes and weights.}
    \label{fig:objects}
    \vspace{-0.2cm}
\end{figure*}

%% file: segments/experiment-setup.tex
\begin{figure}[t]
    \centering
    \includegraphics[scale=0.5]{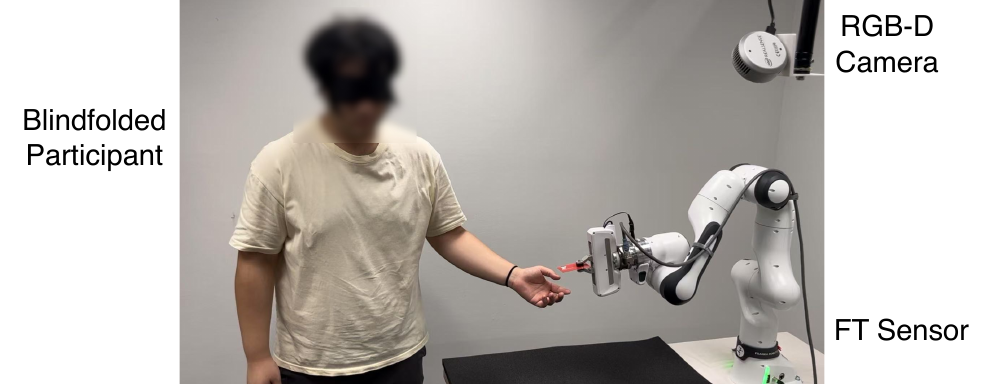}
    \vspace{-0.2cm}
    \caption{Hardware setup. The ATI Gamma sensor and Franka Hand are mounted on a Franka Research 3 arm, and an Intel Realsense L515 RGB-D camera is mounted at the top.}
    \label{fig:experiment-setup}
\end{figure}

%% file: segments/objective-results.tex
\begin{figure}[t]
    \centering
    \includegraphics[scale=0.5]{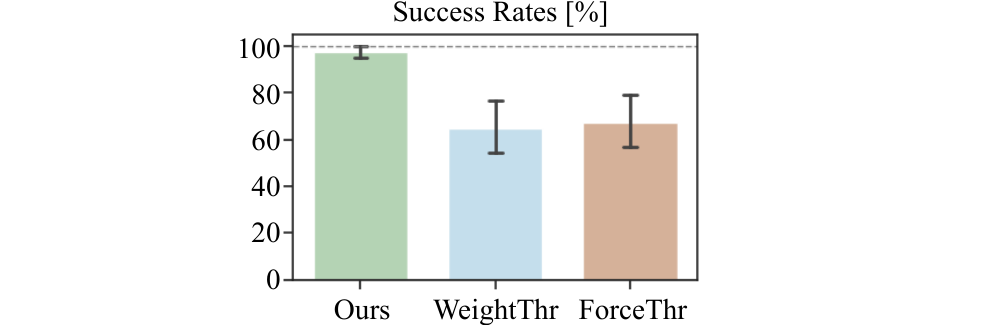}
    \vspace{-0.2cm}
    \caption{Success rates for each method across 12 participants; error bars indicate 95\% confidence intervals. Our method significantly outperforms the baselines.}
    \label{fig:objective-results}
\end{figure}

%% file: segments/survey-scores.tex
\begin{table}[t]
\centering
\caption{RTLX and SUS Survey Scores (0-100)}
\begin{tabular}{lccc}
\toprule
Metric & Ours & \baselineForce{} & \baselineWeight{} \\
\midrule
Raw NASA TLX (RTLX) $\downarrow$ & 34.6 & \textbf{28.9} & 39.3 \\
System Usability Scale (SUS) $\uparrow$ & 63.5 & \textbf{67.9} & 56.5 \\
\bottomrule
\end{tabular}\label{tab:survey-scores}
\end{table}

%% file: references.bib
@string{IJRR = "Int. J. Robotics Research"}

@string{RAL = "IEEE Robotics \& Automation Letters"}

@string{JHRI = "Journal of Human-Robot Interaction"}

@string{T-RO = "IEEE Trans. on Robotics"}

@string{T-ASE = "IEEE Trans. on Automatic Science and Engineering"}

@string{T-CST = "IEEE Trans. on Control System Technology"}

@string{HRI = "Proc. ACM/IEEE Int. Conf. on Human-Robot Interaction"}

@string{Humanoids = "Proc. IEEE-RAS Int. Conf. on Humanoid Robots"}

@string{ICRA = "Proc. IEEE Int. Conf. on Robotics \& Automation"}

@string{IROS = "Proc. IEEE/RSJ Int. Conf. on Intelligent Robots \& Systems"}

@string{ISRR = "Proc. Int. Symp. on Robotics Research"}

@string{ICRR = "Proc. IEEE Int. Conf. on Rehabilitation Robotics"}

@article{ortenzi2021object_handovers,
	title        = {Object Handovers: A Review for Robotics},
	author       = {Ortenzi, Valerio and Cosgun, Akansel and Pardi, Tommaso and Chan, Wesley P. and Croft, Elizabeth and Kulić, Dana},
	year         = 2021,
	journal      = T-RO,
	volume       = {},
	number       = {},
	pages        = {1--19}
}

@inproceedings{bohren2011towards,
	title        = {Towards Autonomous Robotic Butlers: Lessons Learned with the {PR2}},
	author       = {Bohren, Jonathan and Rusu, Radu Bogdan and Gil Jones, E. and Marder-Eppstein, Eitan and Pantofaru, Caroline and Wise, Melonee and Mösenlechner, Lorenz and Meeussen, Wim and Holzer, Stefan},
	year         = 2011,
	booktitle    = ICRA
}

@inproceedings{koene2014experimental,
	title        = {Experimental Testing of The {CogLaboration} Prototype System For Fluent Human-Robot Object Handover Interactions},
	author       = {Ansgar Koene and Satoshi Endo and Anthony Remazeilles and Miguel Prada and Alan M. Wing},
	year         = 2014,
	booktitle    = {{IEEE} Int. Symp. on Robot and Human Interactive Communication}
}

@article{wagner2024robotic,
	title        = {Robotic Scrub Nurse To Anticipate Surgical Instruments Based on Real-Time Laparoscopic Video Analysis},
	author       = {Wagner, Lars and Jourdan, Sara and Mayer, Leon and M{\"u}ller, Carolin and Bernhard, Lukas and Kolb, Sven and Harb, Farid and Jell, Alissa and Berlet, Maximilian and Feussner, Hubertus and others},
	year         = 2024,
	journal      = {Communications Medicine},
	volume       = 4,
	number       = 1,
	pages        = 156
}

@inproceedings{ferrari2024compliant,
	title        = {Compliant Blind Handover Control for Human-Robot Collaboration},
	author       = {Ferrari, Davide and Pupa, Andrea and Secchi, Cristian},
	year         = 2024,
	booktitle    = IROS
}

@article{strabala2013toward,
	title        = {Toward Seamless Human-Robot Handovers},
	author       = {Strabala, Kyle and Lee, Min Kyung and Dragan, Anca and Forlizzi, Jodi and Srinivasa, Siddhartha S. and Cakmak, Maya and Micelli, Vincenzo},
	year         = 2013,
	month        = {feb},
	journal      = JHRI,
	volume       = 2,
	number       = 1,
	pages        = {112–132},
	issue_date   = {February 2013},
	numpages     = 21
}

@article{deyle2010rfid,
	title        = {{RFID}-Guided Robots for Pervasive Automation},
	author       = {Deyle, Travis and Nguyen, Hai and Reynolds, Matthew and Kemp, Charlie},
	year         = 2010,
	journal      = {{IEEE Pervasive Computing}},
	volume       = 9,
	number       = 2,
	pages        = {37--45}
}

@inproceedings{eguiluz2017reliable,
	title        = {Reliable Object Handover Through Tactile Force Sensing And Effort Control in The Shadow Robot Hand},
	author       = {Eguíluz, A. Gómez and Rañó, I. and Coleman, S. A. and McGinnity, T. M.},
	year         = 2017,
	booktitle    = ICRA
}

@inproceedings{han2019effects,
	title        = {The Effects of Proactive Release Behaviors During Human-Robot Handovers},
	author       = {Han, Zhao and Yanco, Holly},
	year         = 2019,
	booktitle    = HRI
}

@inproceedings{nagata1998delivery,
	title        = {Delivery by Hand Between Human and Robot Based on Fingertip Force-Torque Information},
	author       = {Nagata, K. and Oosaki, Y. and Kakikura, M. and Tsukune, H.},
	year         = 1998,
	booktitle    = IROS
}

@article{he2022bidirectional,
	title        = {Bidirectional Human-Robot Bimanual Handover of Big Planar Object With Vertical Posture},
	author       = {He, Wei and Li, Jiashu and Yan, Zichen and Chen, Fei},
	year         = 2022,
	journal      = T-ASE,
	volume       = 19,
	number       = 2,
	pages        = {1180--1191}
}

@article{chan2013human,
	title        = {A Human-Inspired Object Handover Controller},
	author       = {Wesley P Chan and Chris AC Parker and HF Machiel Van der Loos and Elizabeth A Croft},
	year         = 2013,
	journal      = IJRR,
	volume       = 32,
	number       = 8,
	pages        = {971--983}
}

@inproceedings{medina2016human,
	title        = {A Human-Inspired Controller for Fluid Human-Robot Handovers},
	author       = {Medina, Jos{\'e} R and Duvallet, Felix and Karnam, Murali and Billard, Aude},
	year         = 2016,
	booktitle    = Humanoids,
	pages        = {324--331}
}

@inproceedings{kupcsik2018learning,
	title        = {Learning Dynamic Robot-to-Human Object Handover from Human Feedback},
	author       = {Kupcsik, Andras and Hsu, David and Lee, Wee Sun},
	year         = 2018,
	booktitle    = ISRR
}

@inproceedings{parastegari2016fail,
	title        = {A Fail-Safe Object Handover Controller},
	author       = {Parastegari, Sina and Noohi, Ehsan and Abbasi, Bahareh and Žefran, Miloš},
	year         = 2016,
	booktitle    = ICRA
}

@inproceedings{grigore2013joint,
	title        = {Joint Action Understanding Improves Robot-to-Human Object Handover},
	author       = {Elena Corina Grigore and Kerstin Eder and Anthony G. Pipe and Chris Melhuish and Ute Leonards},
	year         = 2013,
	booktitle    = IROS
}

@inproceedings{khanna2022human,
	title        = {Human Inspired Grip-Release Technique for Robot-Human Handovers},
	author       = {Khanna, Parag and Björkman, Mårten and Smith, Christian},
	year         = 2022,
	booktitle    = Humanoids
}

@inproceedings{love1995environment,
	title        = {Environment Estimation for Enhanced Impedance Control},
	author       = {Love, L.J. and Book, W.J.},
	year         = 1995,
	booktitle    = ICRA
}

@article{erickson2003contact,
	title        = {Contact Stiffness and Damping Estimation for Robotic Systems},
	author       = {D. Erickson and M. Weber and I. Sharf},
	year         = 2003,
	journal      = IJRR,
	volume       = 22,
	number       = 1,
	pages        = {41--57}
}

@article{diolaiti2005contact,
	title        = {Contact Impedance Estimation for Robotic Systems},
	author       = {Diolaiti, N. and Melchiorri, C. and Stramigioli, S.},
	year         = 2005,
	journal      = T-RO,
	volume       = 21,
	number       = 5,
	pages        = {925--935}
}

@inproceedings{haddadi2008new,
	title        = {A New Method for Online Parameter Estimation of Hunt-Crossley Environment Dynamic Models},
	author       = {Haddadi, Amir and Hashtrudi-Zaad, Keyvan},
	year         = 2008,
	booktitle    = IROS,
	volume       = {},
	number       = {},
	pages        = {981--986}
}

@article{rozo2016learning,
	title        = {Learning Physical Collaborative Robot Behaviors From Human Demonstrations},
	author       = {Rozo, Leonel and Calinon, Sylvain and Caldwell, Darwin G. and Jiménez, Pablo and Torras, Carme},
	year         = 2016,
	journal      = T-RO,
	volume       = 32,
	number       = 3,
	pages        = {513--527}
}

@article{roveda2022sensorless,
	title        = {Sensorless Optimal Interaction Control Exploiting Environment Stiffness Estimation},
	author       = {Roveda, Loris and Shahid, Asad Ali and Iannacci, Niccolò and Piga, Dario},
	year         = 2022,
	journal      = T-CST,
	volume       = 30,
	number       = 1,
	pages        = {218--233}
}

@article{kronander2016passive,
	title        = {Passive Interaction Control With Dynamical Systems},
	author       = {Kronander, Klas and Billard, Aude},
	year         = 2016,
	journal      = RAL,
	volume       = 1,
	number       = 1,
	pages        = {106--113}
}

@inproceedings{luca2005sensorless,
	title        = {Sensorless Robot Collision Detection and Hybrid Force/Motion Control},
	author       = {de Luca, A. and Mattone, R.},
	year         = 2005,
	booktitle    = ICRA
}

@book{bishop2006pattern,
	title        = {Pattern Recognition and Machine Learning},
	author       = {Bishop, Christopher M and Nasrabadi, Nasser M},
	year         = 2006,
}

@book{ben-tal2009robust_optimization,
	title        = {Robust Optimization},
	author       = {Ben-Tal, Aharon and El Ghaoui, Laurent and Nemirovski, Arkadi},
	year         = 2009
}

@book{boyd2004convex_optimization,
	title        = {Convex Optimization},
	author       = {Boyd, Stephen and Boyd, Stephen P and Vandenberghe, Lieven},
	year         = 2004
}

@book{thrun2005probabilistic_robotics,
	title        = {Probabilistic Robotics},
	author       = {Sebastian Thrun and Wolfram Burgard and Dieter Fox},
	year         = 2005
}

@inproceedings{hart2006nasa,
	title        = {{NASA}-task load index ({NASA-TLX}); 20 years later},
	author       = {Hart, Sandra G},
	year         = 2006,
	booktitle    = {Proceedings of the human factors and ergonomics society annual meeting},
}

@article{brooke1996sus,
	title        = {{SUS}-A quick and dirty usability scale},
	author       = {Brooke, John and others},
	year         = 1996,
	journal      = {Usability evaluation in industry},
	volume       = 189,
	number       = 194,
	pages        = {4--7}
}

@article{landis1977measurement,
	title        = {The measurement of observer agreement for categorical data},
	author       = {Landis, J Richard and Koch, Gary G},
	year         = 1977,
	journal      = {Biometrics},
	pages        = {159--174}
}

@inproceedings{choi2009list,
	title        = {A list of household objects for robotic retrieval prioritized by people with ALS},
	author       = {Choi, Young Sang and Deyle, Travis and Chen, Tiffany and Glass, Jonathan D. and Kemp, Charles C.},
	year         = 2009,
	booktitle    = ICRR
}

@inproceedings{calli2015ycb,
	title        = {The {YCB} object and Model set: Towards common benchmarks for manipulation research},
	author       = {Calli, Berk and Singh, Arjun and Walsman, Aaron and Srinivasa, Siddhartha and Abbeel, Pieter and Dollar, Aaron M.},
	year         = 2015,
	booktitle    = ICRA
}

@book{lynch2017modern,
	title = {Modern Robotics},
	author = {Lynch, Kevin M and Park, Frank C},
	year = {2017}
}
